\pdfoutput=1

\documentclass[11pt]{article}

\usepackage{ACL2023}

\usepackage{times}
\usepackage{latexsym}

\usepackage[T1]{fontenc}
\usepackage[utf8]{inputenc}

\usepackage{inconsolata}
\usepackage{multirow}
\usepackage{graphicx}
\usepackage{booktabs}
\usepackage{amsmath}
\usepackage{CJKutf8}
\usepackage{array}
\usepackage{float}
\usepackage{url}
\usepackage{bm}

\usepackage{xpinyin}
\usepackage{enumerate}

\usepackage{microtype}

\usepackage{inconsolata}
\usepackage{graphicx} 
\usepackage{float} 
\usepackage{subcaption}
\usepackage{algorithm}
\usepackage{algpseudocode}
\usepackage{amsthm,amsmath,amssymb}
\usepackage{mathrsfs}

\title{CBSiMT: Mitigating Hallucination in Simultaneous Machine Translation \\ with Weighted Prefix-to-Prefix Training}

\def\first{$^1$}
\def\second{$^2$}

\def\star{$^*$}
\author{Mengge Liu\first\thanks{\ \ The work was done during the author’s internship at Xiaomi.} ~~~ Wen Zhang\second ~~~ Xiang Li\second ~~~ Yanzhi Tian\first\star \\  ~~~  \textbf{Yuhang Guo\first ~~~ Jian Luan\second ~~~ Bin Wang\second ~~~ Shuoying Chen\first}
\\
{ \first {Beijing Institute of Technology, Beijing, China}} \\
{ \second {Xiaomi AI Lab, Beijing, China}} \\ 
}

\begin{document}
\maketitle
\begin{abstract}
Simultaneous machine translation (SiMT) is a challenging task that requires starting translation before the full source sentence is available. Prefix-to-prefix framework is often applied to SiMT, which learns to predict target tokens using only a partial source prefix. However, due to the word order difference between languages, misaligned prefix pairs would make SiMT models suffer from serious hallucination problems, i.e. target outputs that are unfaithful to source inputs.  
Such problems can not only produce target tokens that are not supported by the source prefix, but also hinder generating the correct translation by receiving more source words.
In this work, we propose a {\bf C}onfidence-{\bf B}ased {\bf Si}multaneous {\bf M}achine {\bf T}ranslation (CBSiMT) framework, which uses model confidence to perceive hallucination tokens and mitigates their negative impact with weighted prefix-to-prefix training.
Specifically, token-level and sentence-level weights are calculated based on model confidence and acted on the loss function. We explicitly quantify the faithfulness of the generated target tokens using the token-level weight, and employ the sentence-level weight to alleviate the disturbance of sentence pairs with serious word order differences on the model.
Experimental results on MuST-C English$\Rightarrow$Chinese and WMT15 German$\Rightarrow$English SiMT tasks demonstrate that our method can consistently improve translation quality at most latency regimes, with up to $2$ BLEU scores improvement at low latency.

\end{abstract}
\section{Introduction}
Simultaneous machine translation (SiMT)~\cite{ma2018stacl,ma2020monotonic,miao2021generative,zhang2022information} is generally based on the prefix-to-prefix framework and formalized as a sequence of READ/WRITE actions~\cite{gu2017learning}.
Unlike full-sentence Neural Machine Translation (named offline NMT), SiMT starts translation before the full source sentence is read, which means that the model needs to be learned on the prefix pairs.

However, the training corpus usually contains vast parallel sentence pairs with different word orders, which will produce misaligned source and target prefixes.
We call a sentence pair with different word orders a \emph{non-monotonic sentence pair}, and vice versa.
Once exposed to non-monotonic sentence pairs, SiMT models probably learn to predict the target token when the corresponding source information has not been read yet, which is $\emph{anticipation}$~\cite{ma2018stacl}. This problem puts the model at risk of generating tokens that are completely unrelated to the source prefix, which is called $\emph{hallucination}$~\cite{lee2018hallucinations, muller-etal-2020-domain1} in the MT task.
Hallucinations in the SiMT model may cause error propagation at inference and further affect the overall translation quality.
Thus, non-monotonic sentence pairs in the training corpus seriously interfere with the training of SiMT models.
In this work, we first conduct two preliminary studies, which reveal that: 1) the decrease of the training data monotonicity leads to the performance degradation of the SiMT model and the aggravation of hallucinations; 2) model confidence~\cite{confidence-estimation} plays a crucial role in perceiving hallucination problems in SiMT.
Based on the above two observations, we propose a {\bf C}onfidence-{\bf B}ased {\bf Si}mul{\bf MT} (CBSiMT) framework that utilizes confidence-based weighted prefix-to-prefix training to perceive hallucination tokens and alleviate the negative impact of hallucination tokens on the SiMT model.
Concretely, our model falls into a general prefix-to-prefix framework, consisting of a unidirectional encoder~\cite{elbayad2020efficient} with masked self-attention, 
the cross-attention between the target sequence and the hidden sequence corresponding to each source prefix is computed to acquire all the confidence values for predicting each target token given any source prefix.
We calculate token-level and sentence-level weights at the same time: 1) the token-level weight estimated by confidences is used to reduce the influence of hallucination tokens in the target sequence; 2) the sentence-level weight is used to alleviate the disturbance of sentence pairs with large word order differences. Both weights are applied to the loss function for weighted prefix-to-prefix training.
Experimental results and in-depth analysis on two benchmarks of MuST-C English$\Rightarrow$Chinese (En$\Rightarrow$Zh) and WMT15 German$\Rightarrow$English (De$\Rightarrow$En) show that our proposed framework can effectively alleviate the hallucination prediction problem of the SiMT model, and improve the translation quality at most latency regimes compared to wait-$k$ baseline, with up to $2$ BLEU scores improvement at low latency.

\section{Preliminary Study}
\subsection{Simultaneous Machine Translation}
The SiMT model generates target tokens while receiving source tokens. Assume a sentence pair in the training set consists of the source sentence $\textbf{x}=\left\{x_1,\cdots,x_j,\cdots,x_J\right\}$ and the observed translation $\textbf{y}^{*}=\left\{y_1^{*},\cdots,y_i^{*},\cdots,y_I^{*}\right\}$.
The SiMT model is optimized by minimizing the loss:
\begin{gather} \label{eq:ce-objectives}
    L_{CE} = - \sum\nolimits_{i=1}^I \log P\left(y_i^{*}|\textbf{x}_{\leq g_i},\textbf{y}_{< i};\theta\right)
\end{gather}
where $\theta$ represents the model parameters, $\textbf{x}_{\leq g_i}$ is the source prefix containing the first $g_i$ source tokens, $\textbf{y}_{< i}^{*}$ means the previously generated tokens, and $P(y_i^*|\textbf{x}_{\leq g_i},\textbf{y}_{< i};\theta)$ is the prediction probability of the ground truth token $y_i^{*}$.
\subsection{Effect of Monotonicity on SiMT}
\begin{figure}[!t]
	\centering
	\begin{minipage}[b]{0.46\textwidth}
		\centering
		\includegraphics[width=\textwidth]{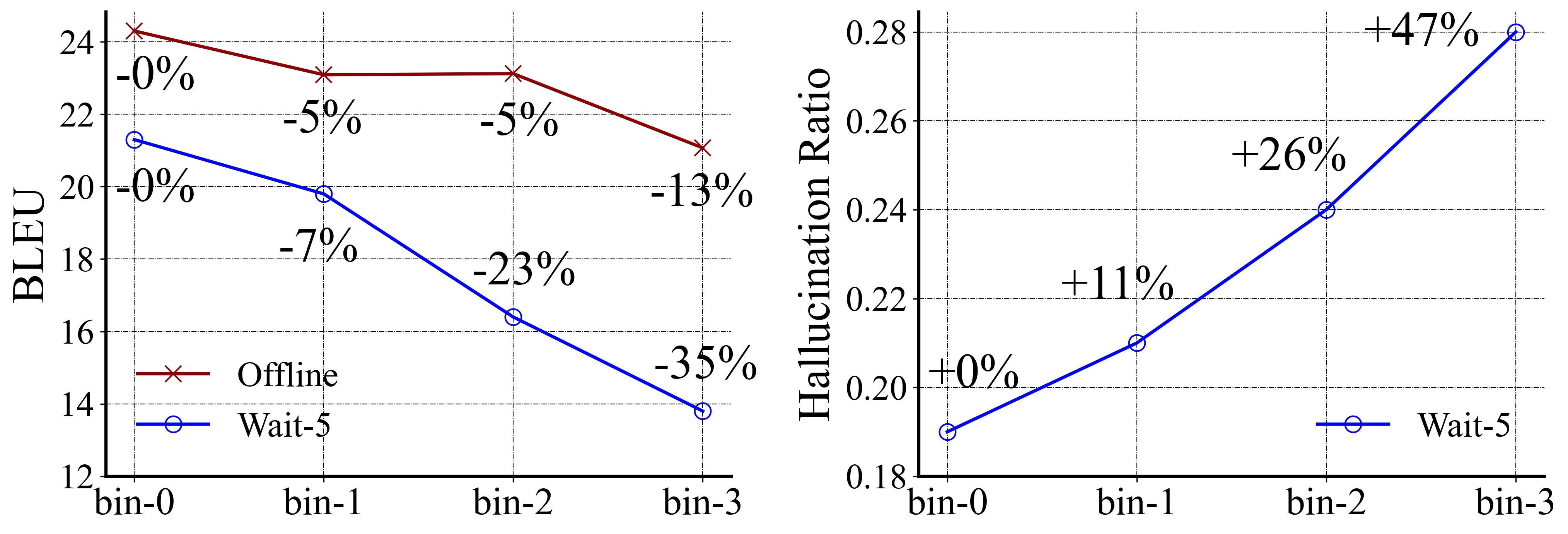}
		\subcaption{CWMT En$\Rightarrow$Zh}
		\label{fig-monoana-cwmt}
	\end{minipage}
	\begin{minipage}[b]{0.46\textwidth}
		\centering
		\includegraphics[width=\textwidth]{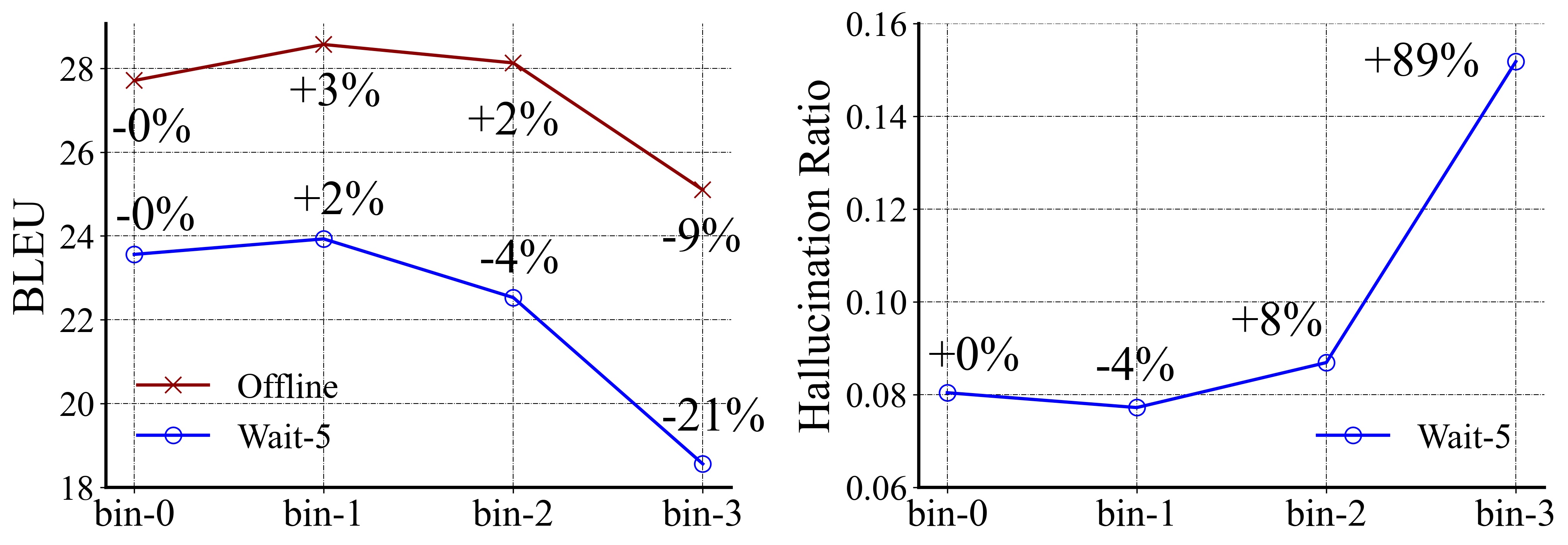}
		\subcaption{WMT15 De$\Rightarrow$En}
		\label{fig-monoana-wmt}
	\end{minipage} 
	\caption{Effect of the non-monotonic sentence pairs on the SiMT model. The bins are sorted in descending order of monotonicity, with $\mathrm{bin}$-$0$ and $\mathrm{bin}$-$3$ having the best and worst monotonicity respectively. The hallucination ratio denotes the proportion of hallucinated tokens in the generated sentences at inference.}
	\label{figs:monotonic-analysis}
 \vspace{-0.5cm}
\end{figure}
We design an experiment to explore the effect of training data monotonicity on the performance of SiMT models and hallucination issues. Similar to the $k$-anticipation rate (${AR}_k$) defined by~\citet{chen2020improving}, we define a metric called Average Anticipation (abbreviated as AA) to evaluate the word order difference, which is computed by:
$AA = \frac{1}{n} \sum_{\tiny{(x_i,y_j)} \in \tiny{As}} \max\left(i-j, 0\right)$, where $As$ represents the set of word alignments, $\tiny{(x_i,y_j)} \in \tiny{As}$ means that the $i^{th}$ token in the source sentence is aligned with the $j^{th}$ token in the target sentence, $n$ is the number of alignment pairs, and the smaller the AA value, the better the monotonicity of the sentence pair.
We sort the sentence pairs in the training set of CWMT En$\Rightarrow$Zh and WMT15 De$\Rightarrow$En in ascending order according to the AA value, and then divide each training set equally into $4$ bins, and $\mathrm{bin}$-$0$ has the best monotonicity.
The offline and wait-$k$~($k$=$5$) models are trained separately on each bin and evaluated on the same test set. We apply beam search with a beam size of $5$ and wait-$5$ policy to the offline and wait-$k$ models at inference.
For each model, we present the trend of the decrease rate of the BLEU score on each bin relative to the score on $\mathrm{bin}$-$0$. For wait-$k$ models, we also present the trend of hallucination ratio~(details are described in Sec.~\ref{sec:hallucination-ratio}).
As shown in Figure~\ref{figs:monotonic-analysis}, no matter for the CWMT task or the WMT task, as the monotonicity of the training data decreases, the performance decline of the wait-$k$ model is more significant than that of the offline model, accompanied by a massive increase in the hallucination rate.
In summary, the SiMT model is sensitive to non-monotonic sentence pairs, which could result in serious hallucination phenomena and severely degrades the translation quality of the model.
\begin{figure}[t!] 
\centering 
\includegraphics[width=0.46\textwidth]{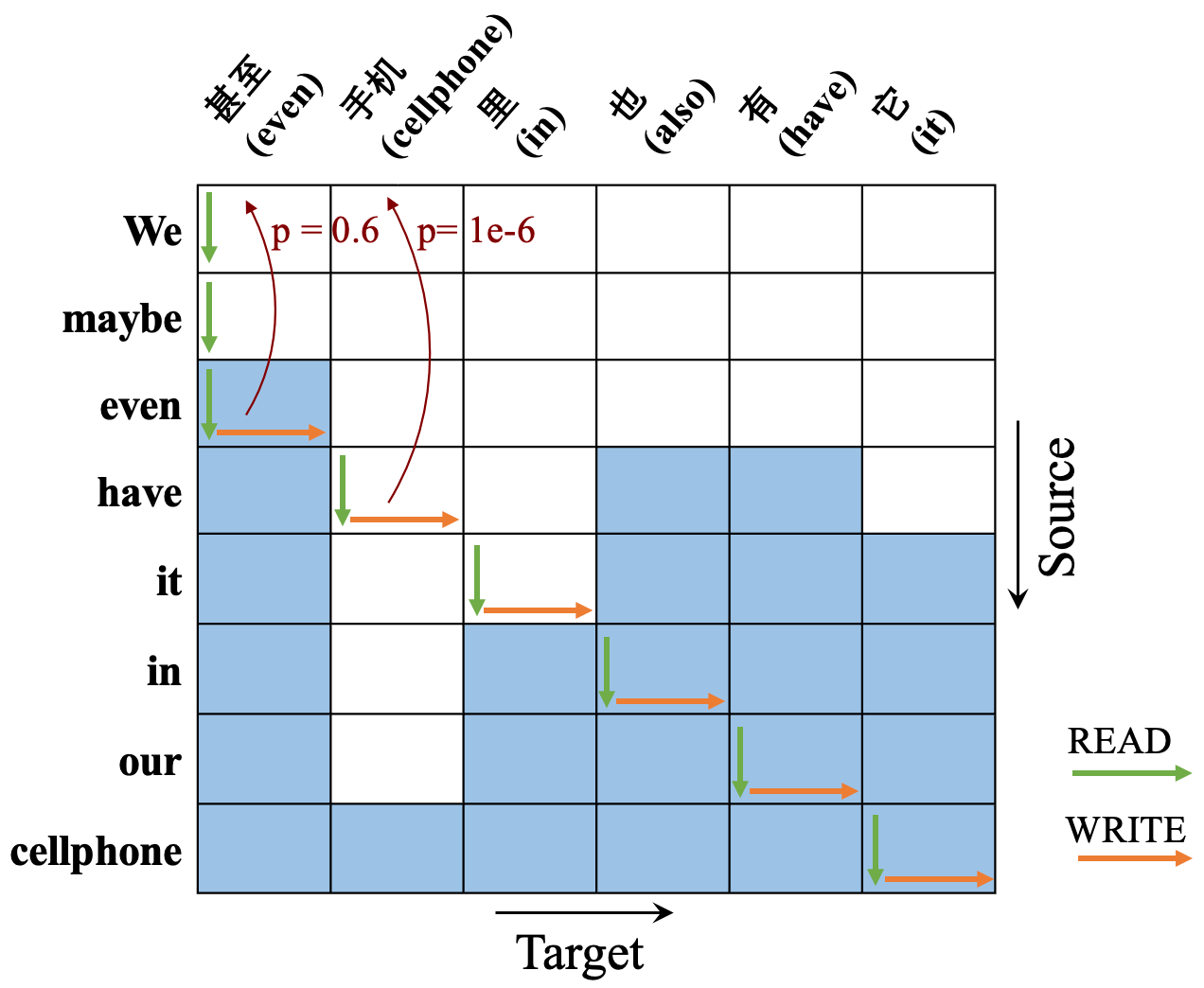} 
\caption{Diagram of model confidence in SiMT model. The wait-$3$ READ/WRITE policy is simulated by arrows. Force decoding is performed for each possible WRITE action. Positions in the matrix with confidence greater than $1e$-$3$ are marked in blue.} 
\label{fig:confidence_ana} 
\vspace{-0.3cm}
\end{figure}
\subsection{Confidence and Hallucinations}
MT community has long been aware of the importance of model confidence, which is usually adopted to measure the success or failure of model predictions~\citep{wang2019improving,kumar2019calibration,wang2020inference,wan2020self,liu2021confidence,zhou2022confidence,lu2022learning}.
\begin{CJK*}{UTF8}{gbsn}
To explore the correlation between confidence and hallucinations, we apply forced decoding to ground truth sequences based on an offline NMT model trained on the MuST-C En$\Rightarrow$Zh training set. When the $j$-th source word $x_j$ is read, we force the offline model to predict all ground truth tokens, i.e. pick the prediction probability corresponding to the ground truth tokens to estimate the confidence, and put them into the $j$-th row of the confidence matrix, as shown in Figure~\ref{fig:confidence_ana}.
Among them, the $i$-th column represents the confidence of the $i$-th ground truth token being predicted after all source prefixes of different lengths are read.
We use an example to simulate the wait-$3$ inference policy, and illustrate the corresponding READ/WRITE actions in Figure~\ref{fig:confidence_ana}. Given $3$ source words, the target word ``甚至(even)'' is predicted with high confidence. On the contrary, the next predicted word ``手机(cellphone)'' has extremely low confidence since the corresponding source word ``cellphone'' has not been read yet. Therefore, the generated target word ``手机(cellphone)'' can be regarded as a hallucination token.
The above observations prove that confidence in the SiMT model is a key metric that can reflect hallucinations, which has also been confirmed by~\citet{zheng2020simultaneous}.
\end{CJK*}
\begin{figure*}[t!] 
\centering 
\includegraphics[scale=0.24]{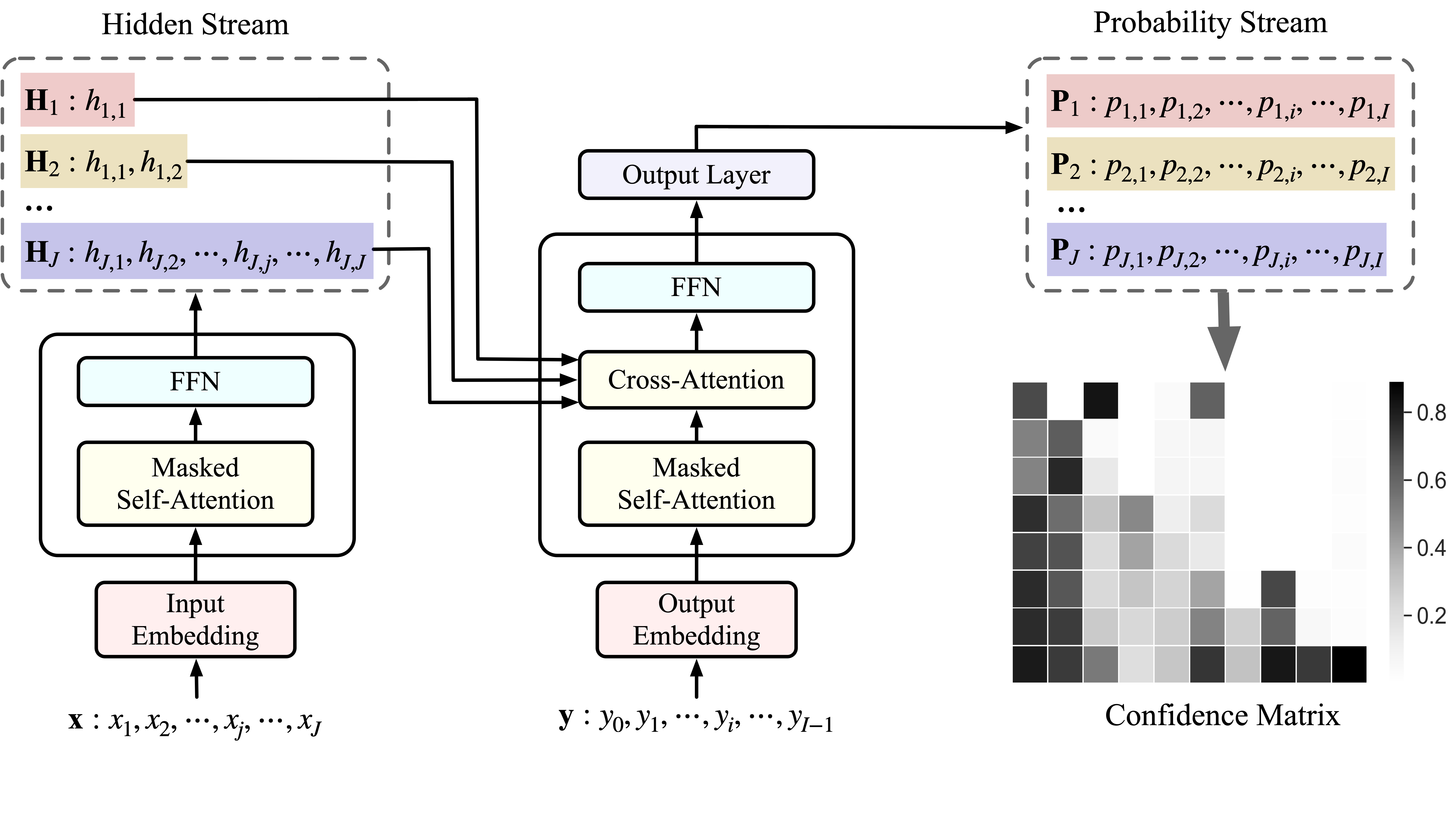} 
\vspace{-0.9cm}
\caption{Illustration of the CBSiMT framework.} 
\label{fig:streaming-transformer} 
\end{figure*}
\section{Confidence-Based SiMT Framework}
\subsection{The Overview of CBSiMT}\label{sec:arch}
As shown in Figure~\ref{fig:streaming-transformer}, our proposed CBSiMT falls into the prefix-to-prefix training framework. We adopt Masked Self-Attention in each layer in the CBSiMT encoder, which is similar to the unidirectional encoder in~\citet{elbayad2020efficient}. 
In order to be consistent with the real-time inference process, the encoder will output a $\emph{hidden stream}$, including the hidden state of multiple source prefixes. The decoder will predict target tokens conditioned on each hidden state, and output a $\emph{probability stream}$.
\subsubsection{Hidden Stream}\label{sec:hidden_stream}
Instead of the hidden state sequence corresponding to the full source sentence or the hidden state sequence of a certain source prefix, the decoder will receive hidden state sequences of all prefixes to ensure that the model can take into account all source prefixes (marked as $\emph{hidden stream}$) when predicting the target token.

Given a prefix $\textbf{x}_{\leq j}=\left\{x_1,\cdots,x_j\right\}$ of the full source sentence $\textbf{x}=\left\{x_1,\cdots,x_j,\cdots,x_J\right\}$ as input, the CBSiMT encoder generates a hidden state sequence $\textbf{H}_j=(h_{j,1},\cdots,h_{j,k},\cdots,h_{j,j})$ corresponding to the source prefix $\textbf{x}_{\leq j}$. The hidden state sequence $\textbf{H}_j$ is expressed as the following formula:
\begin{gather} \label{eq:hidden-stream}
    \textbf{H}_j = \text{Encoder}\left(\textbf{x}_{\leq j}\right)
\end{gather}
All the prefixes of the source sentence $\textbf{x}$ pass through the encoder to obtain $\emph{hidden stream}$ $\mathcal{H}=\left\{\textbf{H}_1,\cdots,\textbf{H}_j,\cdots,\textbf{H}_J\right\}$.
Benefiting from Masked Self-Attention, we can first output the hidden state sequence $\textbf{H}_J$ corresponding to the full source sentence $\textbf{x}$ through one forward calculation, and then obtain the hidden state sequence of all prefixes by truncation to improve computational efficiency.

\subsubsection{Probability Stream}\label{sec:adaptive_decoder}
With the hidden stream, SiMT models make predictions based on source prefixes.
Our CBSiMT decoder is aware of source prefixes of different lengths during optimization.
After reading the first $j$ source tokens, the predicted probability $p_{j,i}$ of the $i$-th target token is calculated as follows:
\begin{gather} \label{eq:probility-stream}
    o_{j,i} = \text{Decoder}\left(\textbf{y}_{<i}, \textbf{H}_j\right) \\
    p_{j,i} = \text{Index}\left(\text{Softmax}\left(\text{OutputLayer}\left(o_{j,i}\right)\right)\right)
\end{gather}
where $\textbf{H}_j$ is the encoder output corresponding to the source prefix of length $j$, as shown in Eq.\ref{eq:hidden-stream}. $\textbf{y}_{<i}$ represents the previous $i$-$1$ target tokens and $o_{j,i}$ is the output of CBSiMT decoder.
After $o_{j,i}$ is projected through the output layer and softmax layer, the predicted probability of the ground truth token $y_i^*$ is picked out as $p_{j,i}$ with index operator.
For each hidden state sequence $\textbf{H}_j$ in the hidden stream, the CBSiMT decoder can predict the probability $p_{j,i}$ corresponding to each token $y_i^*$ in the ground truth sequence. The resulting probability matrix is labeled as the $\emph{probability stream}$, as shown in Figure~\ref{fig:streaming-transformer}.
The probability stream contains predicted probabilities based on each source prefix, which also could be viewed as the confidence matrix.
\subsection{Adaptive Training} \label{sec:objectives}
With well-designed token-level and sentence-level weights, we regularize the training objective as follows to make the model dynamically perceive data monotonicity and alleviate the disturbance of non-monotonic sentence pairs on the model:
\begin{gather} \label{eq:objectives}
    L_{CBSiMT} = -\beta \times \frac{1}{J} \sum_{i=1}^I \sum_{j=1}^J \alpha_{j,i} \times \log p_{j,i}
\end{gather}
where $\beta$ is the sentence-level weight for a sentence pair, and $\alpha_{j,i}$ is the token-level weight based on each possible prefix. We elaborate on token-level and sentence-level weights in the next two sections.
\begin{figure*}[htbp]
	\centering
	\begin{minipage}[c]{0.24\textwidth}
		\centering
		\includegraphics[width=\textwidth]{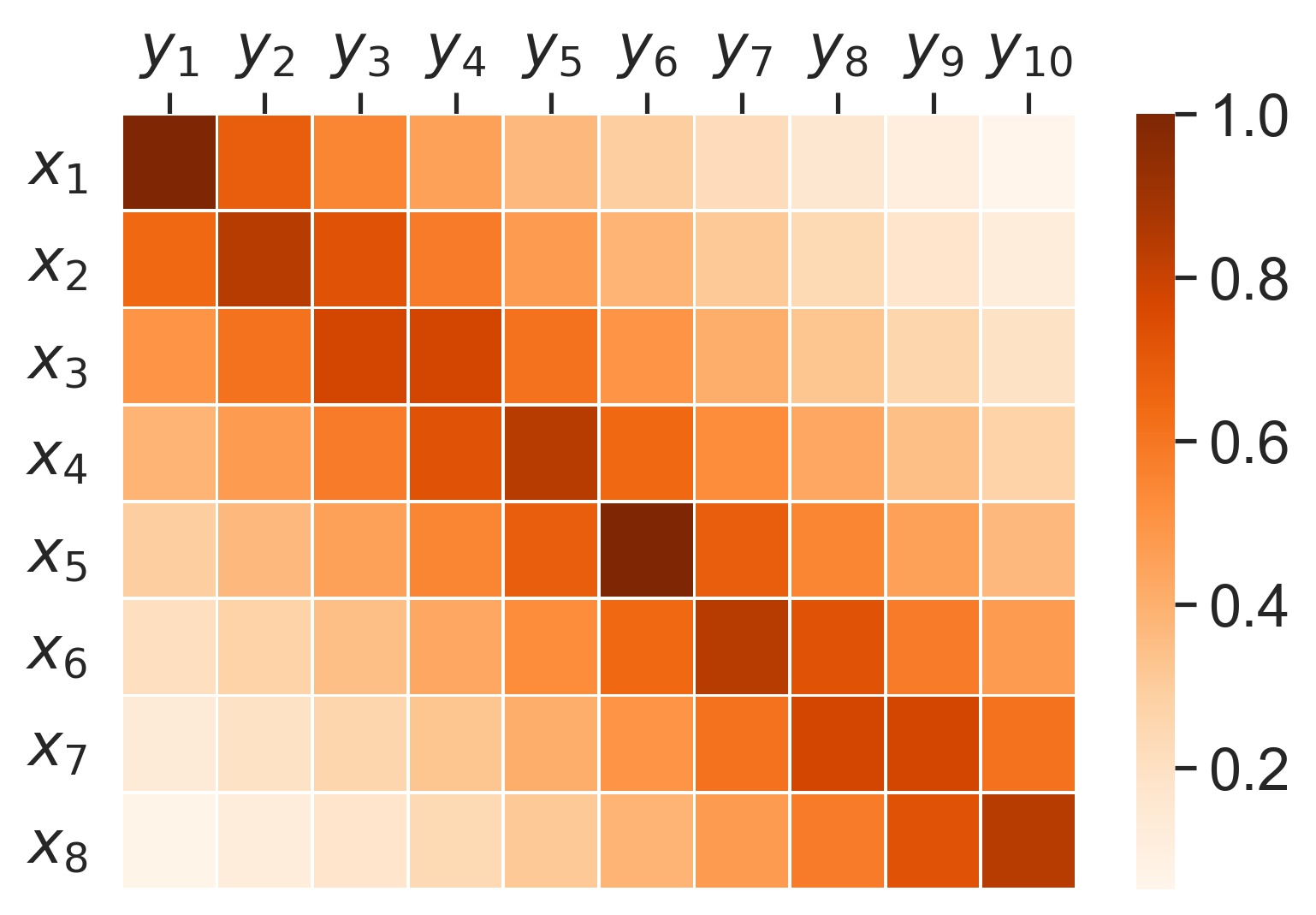}
		\subcaption{$D_{j,i}~(\lambda = 0.5)$}
		\label{fig-lambda0.5}
	\end{minipage}
	\begin{minipage}[c]{0.24\textwidth}
		\centering
		\includegraphics[width=\textwidth]{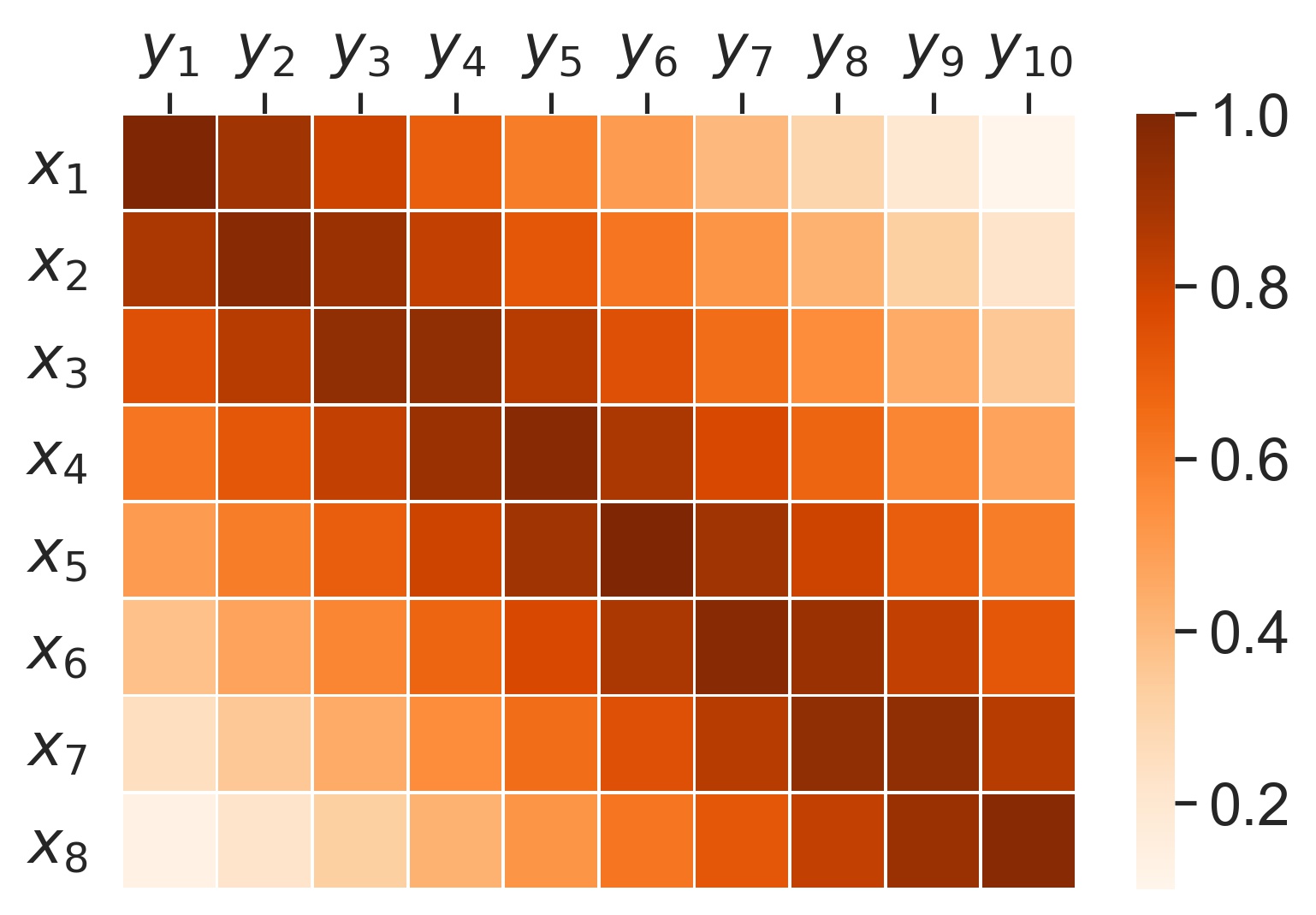}
		\subcaption{$D_{j,i}~(\lambda = 1)$}
		\label{fig-lambda1}
	\end{minipage} 
	\begin{minipage}[c]{0.24\textwidth}
		\centering
		\includegraphics[width=\textwidth]{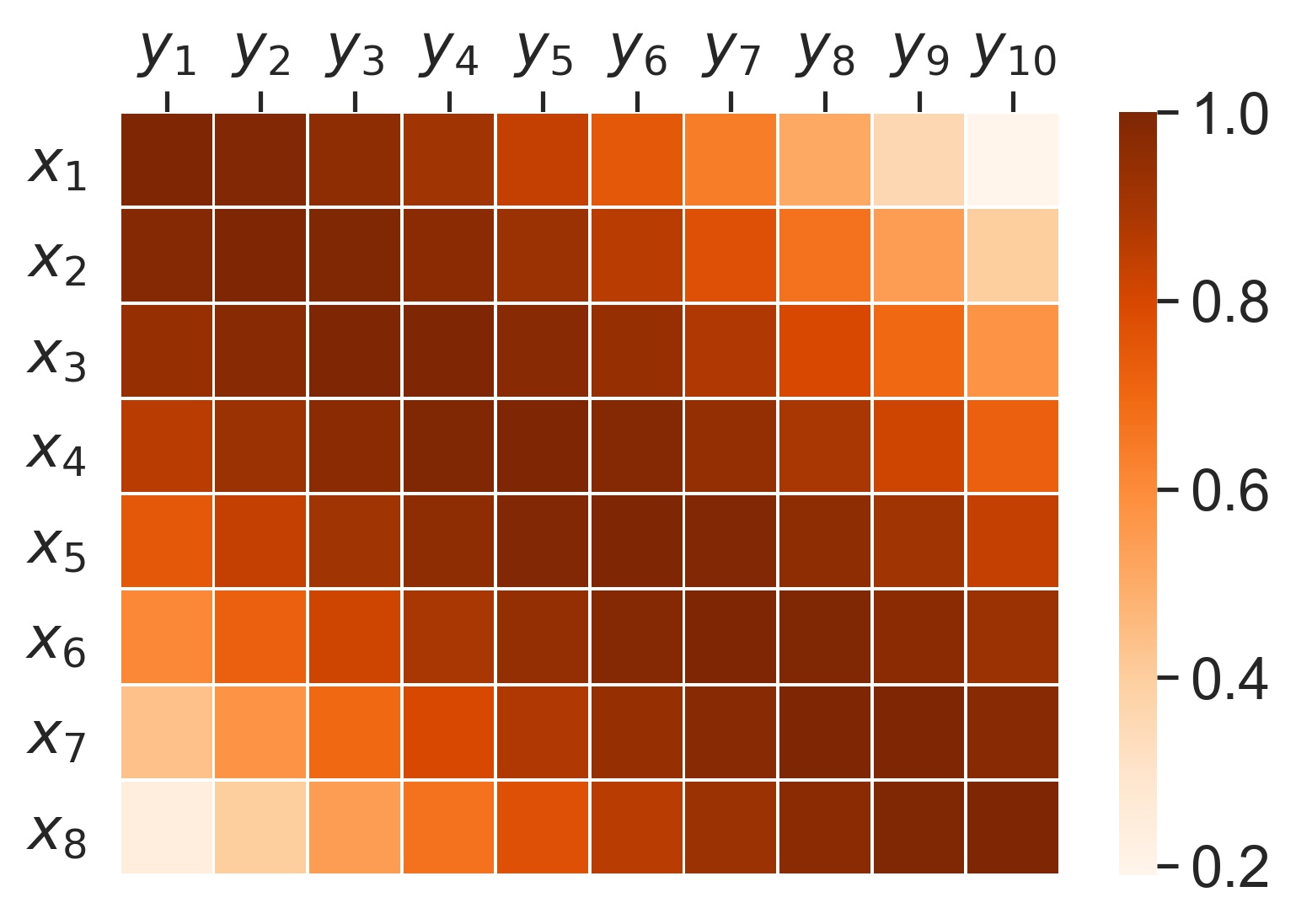}
		\subcaption{$D_{j,i}~(\lambda = 2)$}
		\label{fig-lambda2}
	\end{minipage}
        \begin{minipage}[c]{0.24\textwidth}
		\centering
		\includegraphics[width=\textwidth]{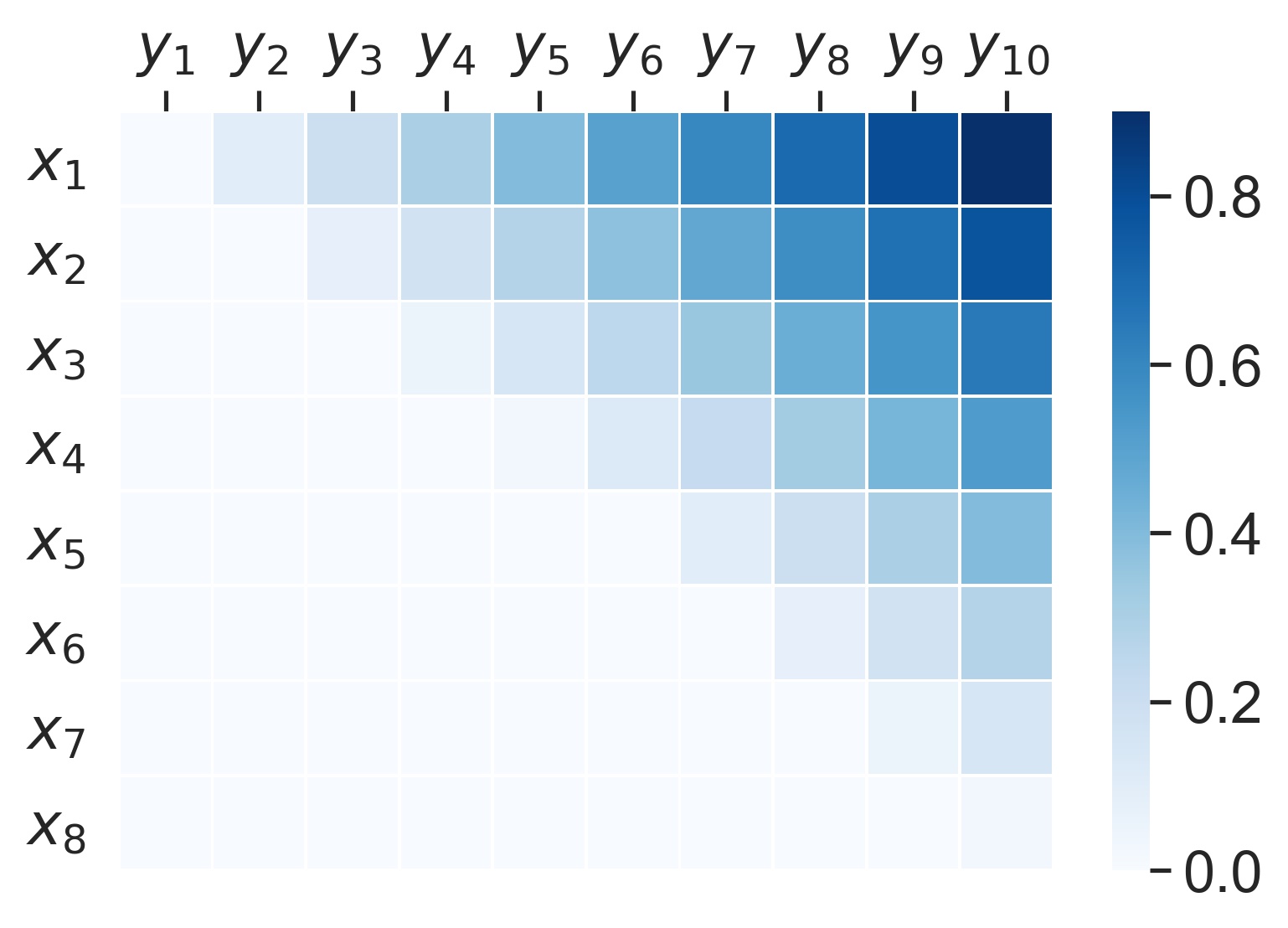}
		\subcaption{$c_{ji}$}
		\label{fig-reordering-cost}
	\end{minipage}
	\caption{Schematic diagram of matrixs used in the training of CBSiMT framework.}
	\label{figs:diagonal-lambda-cost}
    \vspace{-0.5cm}
\end{figure*}

\subsubsection{Token-Level Weight} \label{sec:token_level_weight}
We exactly use token-level predicted probabilities as token-level confidence, which can reflect the correctness of each generated token.
However, there are two main problems in SiMT prediction: 1) $\emph{hallucination}$: the model uses few source tokens to predict the tokens at the end of the target sequence, corresponding to the confidence in the upper right corner of the confidence matrix, marked as $p_{j,i;j\ll i}$;
2) $\emph{high-latency prediction}$: many tokens are read to generate the tokens at the beginning of the target sequence, and the corresponding confidences are located in the lower-left corner of the confidence matrix, denoted as $p_{j,i;j\gg i}$.
Therefore we propose a diagonal regularization method to attenuate the model's predictions in both cases.

Intuitively, predictions close to the diagonal are nearly ideal READ/WRITE paths with $0$ latency, and positions farther from the diagonal should be assigned smaller weights.
For each predicted probability $p_{j,i}$, the token-level weight $\alpha_{j,i}$ is calculated:
\begin{gather} 
 \label{eq:distance} d_{j,i} = |\frac{i}{I} - \frac{j}{J}| \\ 
 \label{eq:diag-regularization} D_{j,i} = 1-d_{j,i}^\lambda \\ 
 \label{eq:token-weight} \alpha_{j,i} =p_{j,i}^\gamma \times D_{j,i}
\end{gather}
where $d_{j,i}$ can be roughly considered as the distance between $p_{j,i}$ and the diagonal in the confidence matrix, hyperparameter $\gamma$ is used for scaling confidence and set $\gamma$ to be $0.25$, and $\lambda$ is used to control the degree of regularization. As shown in Figure~\ref{fig-lambda0.5}\textasciitilde\ref{fig-lambda2}, the smaller the value of $\lambda$, the more concentrated the diagonal regularization, meaning the regularization constraint is stronger.




\subsubsection{Sentence-Level Weight}
The sentence-level weight $\beta$ acts on the loss to alleviate the disturbance of non-monotonic sentence pairs at the sentence level, and is calculated by:
\begin{gather} 
 \label{eq:reorder-cost} c_{j,i} = \max(0, \frac{i}{I} - \frac{j}{J}) \\ 
 \label{eq:reorder-score} C = \sum\nolimits_{i=1}^I \sum\nolimits_{j=1}^J p_{j,i} \times c_{j,i} \\ 
 \label{eq:sentence-weight} \beta = \max\left\{0,1-\left(C - \mu_{batch}\right)/\sigma_{batch}\right\}
\end{gather}
where $c_{j,i}$ means the reordering cost corresponding to each predicted probability $p_{j,i}$, as illustrated in Figure~\ref{fig-reordering-cost}.
Large reordering costs in the upper right corner (corresponding to $p_{j,i;j\ll i}$ mentioned in Sec.~\ref{sec:token_level_weight}) may lead to hallucination prediction.
Then, the reordering score $C$ of a sentence pair is used to measure the word order difference and normalized by the mean $\mu_{batch}$ and standard deviation $\sigma_{batch}$ of the $C$ values in each batch during training.
Sentence pairs with more significant word order differences are more non-monotonic and should have smaller sentence-level weights.
\subsection{Inference Policy} \label{sec:inference}
Consistent with confidence-based training, CBSiMT makes READ/WRITE decisions based on the predicted confidence $p_{pred}$ at each decoding step.
At a certain step, $\mathcal{X}$ and $\mathcal{Y}$ represent the read source prefix and the generated target prefix, respectively.
The expression $\mathcal{X} \gets \mathcal{X} \bullet x$ represents the token $x$ is appended to the end of the sequence $\mathcal{X}$.
At inference, the number of lagging tokens $l \gets |\mathcal{X}| - |\mathcal{Y}|$ is calculated at each step, and the threshold varies with $l$, denoted as $th_l$.
When $l$ is small, the source information may not be sufficient for the model to perform WRITE actions, so $th_l$ needs to be set higher to increase the possibility of performing READ actions and only output very confident tokens, and vice versa. $th_l$ is defined:
$$th_l \gets th_{max} \times\left(1 - \left(l/l_{max}\right)^\delta\right)$$
where $th_{l}$ reaches the maximum when $l$=$0$, as $l$ increases to the maximum $l_{max}$, $th_l$ reaches the minimum $0$, at which point WRITE action must be performed. Hyperparameter $\delta$ is introduced to obtain translations at different delays.

When the confidence exceeds $th_l$, the WRITE action is performed, and we record the number of read source words as the delay and append it to the delay sequence $\mathcal{D}$; otherwise, the next source word is read. The details are shown in Algorithm~\ref{alg:CBSiMT-inference}.
$\mathcal{Y}$ is the final translation, and the delay sequence $\mathcal{D}$ is used for latency computation.
\begin{algorithm}[h]
    \renewcommand{\algorithmicrequire}{\textbf{Input:}}
    \renewcommand{\algorithmicensure}{\textbf{Output:}}
    \caption{Inference policy}
    \label{alg:CBSiMT-inference}
    \begin{algorithmic}[1]
    \Require
      Transformer model trained with CBSiMT framework $M_{CBSiMT}$, delay sequence $\mathcal{D}$, maximal lagging $l_{max}$, maximal threshold $th_{max}$, hyperparameter $\delta$.
    \State \textbf{Initialization} : $\mathcal{X}$ = $[x_1]$, $\mathcal{Y} = [\ ]$, $\mathcal{D}=[\ ]$
    \While{$y_{|\mathcal{Y}|} \neq EOS$}
        \State $l \gets |\mathcal{X}| - |\mathcal{Y}|$ 
        \State $th_l \gets th_{max} \times\left(1 - \left(l/l_{max}\right)^\delta\right)$
        \State $y_{pred}, p_{pred} = M_{CBSiMT}(\mathcal{X}, \mathcal{Y})$
        \If{$p_{pred} \geq th_l$ or $x_{|\mathcal{X}|} = EOS$ } 
            \State $\mathcal{Y} \gets \mathcal{Y}$ $ \bullet\ y_{pred}$ \Comment{WRITE action}
            \State $\mathcal{D} \gets \mathcal{D} \bullet\ |\mathcal{X}|$
            \Comment{delay number}
        \Else 
            \State $\mathcal{X} \gets \mathcal{X}$ $ \bullet\ x_{next}$ \Comment{READ action}
        \EndIf
    \EndWhile
    \Ensure
      Translation $\mathcal{Y}$, delay sequence $\mathcal{D}$
  \end{algorithmic}
\end{algorithm}

\section{Experiments}
\begin{figure*}[!t]
	\centering
	\begin{minipage}[c]{0.43\textwidth}
		\centering
		\includegraphics[width=\textwidth]{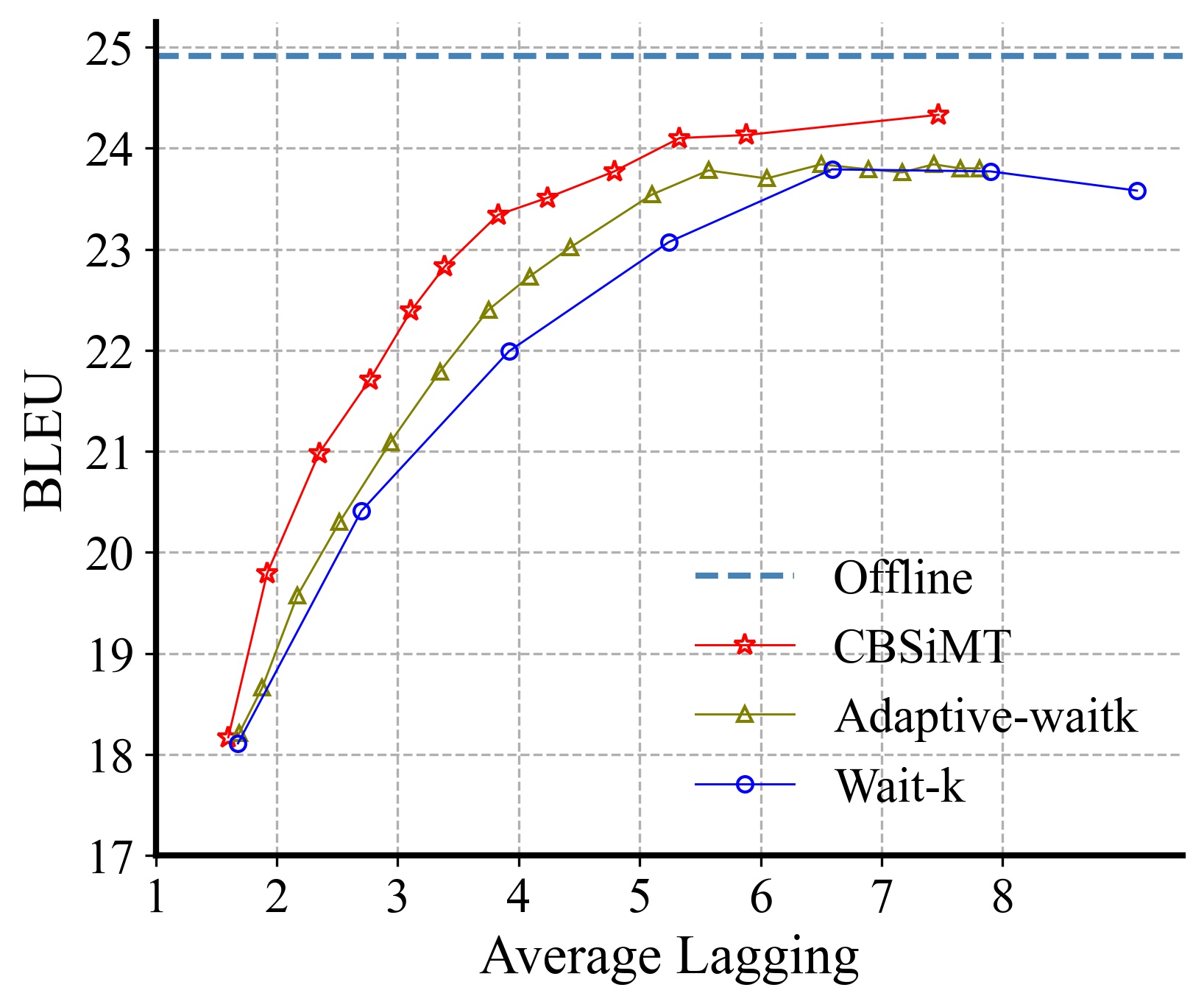}
		\subcaption{MuST-C En$\Rightarrow$Zh}
		\label{fig-mustc}
	\end{minipage}
	\begin{minipage}[c]{0.43\textwidth}
		\centering
		\includegraphics[width=\textwidth]{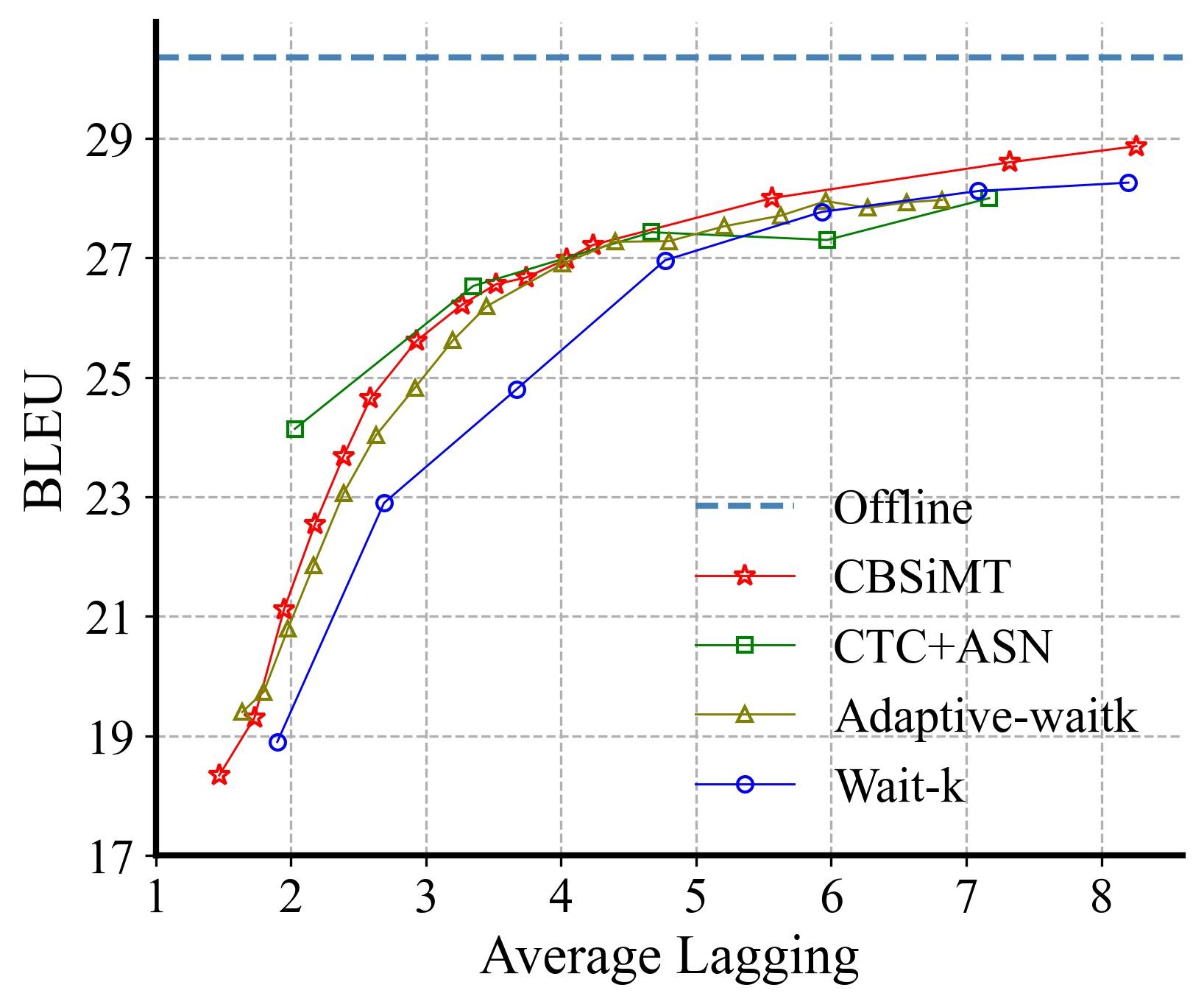}
		\subcaption{WMT15 De$\Rightarrow$En}
		\label{fig-wmt15}
	\end{minipage} 
	\caption{Latency-quality trade off on the MuST-C En$\Rightarrow$Zh and WMT15 De$\Rightarrow$En tasks. Each line corresponds to a simultaneous machine translation system.} \vspace{-0.5cm}
	\label{figs:main-results}
\end{figure*}
\subsection{Datasets}
\paragraph{MuST-C En$\Rightarrow$Zh} We extract the En$\Rightarrow$Zh dataset from MuST-C release v2.0\footnote{\url{https://ict.fbk.eu/must-c-release-v2-0/}}~\citep{di2019must,cattoni2021must}, where the training and validation sets consist of $358,853$ and $1,349$ sentence pairs, respectively. The~\texttt{tst-COMMON} is used as the test set, containing $2841$ pairs.
\paragraph{WMT15 De$\Rightarrow$En}
The training set of WMT15~\footnote{\url{www.statmt.org/wmt15/}} De$\Rightarrow$En datasets contains $4.5$M sentence pairs. We use~\texttt{newstest2013} and~\texttt{newstest2015} as the validation and test sets, respectively, containing $3,000$ and $2,169$ sentence pairs.

~\texttt{Sacremoses}\footnote{\url{https://github.com/alvations/sacremoses}} is employed for English and German tokenization, \texttt{Jieba}\footnote{\url{https://github.com/fxsjy/jieba}} is employed for Chinese word segmentation.
\subsection{Experiments Settings}
To ensure the reliability of the model confidence, we first pre-train an offline NMT model with the full-sentence cross-entropy objective until convergence, and then fine-tune the pre-trained model on the same training set. The learning rate used for pre-training is $2.5e$-$4$. For the MuST-C En$\Rightarrow$Zh and WMT15 De$\Rightarrow$En tasks, we fine-tune the model with the objective function in Equation~\ref{eq:objectives} and set the learning rate to $1e$-$5$ and $5e$-$5$ respectively.
It is worth noting that the hidden sequence of each prefix in each source full sentence needs to be calculated, which may cause memory overflow, especially for long sentences, so we introduce a sampling mechanism to control the size of hidden and probability streams.
In all experiments, at most $10$ source prefixes can be passed through the encoder to generate the hidden stream.
We select hyperparameters and checkpoints according to the performance on the validation set. The following four dominant methods are used as baselines for comparison with CBSiMT:
\begin{list}{\labelitemi}{\leftmargin=1em} \vspace{-0.2cm}
\setlength{\itemsep}{0pt}
\setlength{\parsep}{0pt}
\setlength{\parskip}{0pt}
    \item {\bf Offline}:  Full-sentence Transformer~\citep{vaswani2017attention}, waiting for the full source sentence for translation and applying greedy decoding.
    \item {\bf Wait-$k$}: Wait-$k$~\citep{ma2018stacl} policy, the most widely used fixed policy, which first READ $k$ source tokens, and then alternately performs READ and WRITE actions.  
    \item {\bf Adaptive Wait-$k$}: An heuristic composition of multiple wait-$k$ models~\cite{zheng2020simultaneous} to achieve adaptive inference, which uses prediction probability to make READ/WRITE decisions.
    \item {\bf CTC+ASN}: A framework which decomposes the translation process into the monotonic translation step~(CTC translation) and the reordering step~(auxiliary sorting network, ASN)~\cite{chang2022anticipation}.
\end{list} 

All models are trained with~\texttt{fairseq}~\cite{fairseq-cite} and evaluated with toolkit SimulEval~\footnote{\url{https://github.com/facebookresearch/SimulEval}}~\cite{simuleval-toolkit}.
We use the \emph{transformer\_iwslt\_de\_en} setting for the MuST-C En$\Rightarrow$Zh translation task, and the \emph{transformer\_base} setting for the WMT15 De$\Rightarrow$En translation task, respectively.
All models are optimized by Adam optimizer~\cite{kingma2015adam}.
We report~\texttt{SacreBleu}~\cite{post2018call} as the quality metric and ~\texttt{Average Lagging}~(AL)~\cite{ma2018stacl} as the latency metric in the following experiments. For Chinese and English, latency is calculated at the granularity of characters and tokens respectively.

\subsection{Main Results}
All results are listed in Figure~\ref{figs:main-results}, we compare the performance of the different methods with BLEU-AL curves, and the higher the BLEU score under the same AL regime, the better the method.
Based on the BLEU-AL curves on the validation set, the hyperparameter $\lambda$ is set to $1$ for diagonal regularization on the two benchmarks MuST-C En$\Rightarrow$Zh and WMT15 De$\Rightarrow$En.
At inference, $th_{max}$ is set as $0.9$ to ensure the correct prediction in low latency, $l_{max}$=$9$, $\delta \in \left\{0.1, 0.3, 0.5, 0.9, 1.5, 2.0, 3.0, 5.0 \right\}$ and $l_{max}$=$19$, $\delta \in \left\{ 1.5, 3.0, 5.0\right\}$ are verified to obtain results for different latency regime. 
We implement efficient wait-$k$~\cite{elbayad2020efficient} models as a baseline for SiMT, $k$=$9$ is used for training, and $k$=$\left\{1,3,5,7,9,11,13\right\}$ are used for inference.
The adaptive wait-$k$ policy is implemented with a single wait-$9$ model following the settings in~\citet{zheng2020simultaneous}.
We report the BLEU-AL curves in~\citet{chang2022anticipation} as the result of CTC+ASN model, which utilizes sequence-level distillation~\cite{kim-rush-2016-sequence1} to improve the translation quality.

Our CBSiMT framework achieves the best performance on the MuST-C En$\Rightarrow$Zh task and has a comparable latency-quality trade-off with other adaptive strategies on the WMT15 De$\Rightarrow$En task.
On both tasks, CBSiMT significantly outperforms the fixed wait-$k$ baseline, and can have an advantage of up to $2$ BLEU scores when the AL value falls into the interval [$3$, $4$].
In addition, compared with other dynamic policies, such as the adaptive wait-$k$ policy, CBSiMT still has obvious superiority on two tasks, which indicates that our proposed CBSiMT could produce better READ/WRITE decisions.
With the positive effect brought by sequence-level knowledge distillation, compared with CBSiMT on the De$\Rightarrow$En task, CTC+ASN performs better at low latency, but loses its advantage when the latency value exceeds $5$.
It is worth mentioning that the CBSiMT framework significantly outperforms all other baseline systems on the MuST-C En$\Rightarrow$Zh task. The data of the MuST-C En$\Rightarrow$Zh task is collected from TED talks, so our CBSiMT framework is more suitable for speech scenarios.

\section{Analysis}
We conduct in-depth analyses to explore why CBSiMT improves SiMT and its effect on hallucinations. 
\vspace{-0.3cm}
\subsection{Ablation Studies}
\label{sec:ablation}
\begin{table}[!t]
\centering
\renewcommand\arraystretch{1.1}
{\footnotesize
\begin{tabular}{lcccc}
\toprule
\textbf{Model} & \textbf{BLEU}  & \textbf{$\Delta$} & \textbf{AL} & \textbf{$\Delta$} \\
\hline
\textbf{CBSiMT} & $\textbf{24.61}$ & ~ & $\textbf{3.36}$ & ~ \\ 
\quad - w/o $\beta$ & $24.33$ & -$0.28$ & $3.43$ & +$0.07$ \\ 
\quad - w/o $D_{j,i}$ & $24.30$ & -$0.31$ & $3.45$ & +$0.09$ \\ 
\quad - w/o $\beta,D_{j,i}$ & $24.27$ & -$0.34$ & $3.45$ & +$0.09$ \\ 
\bottomrule
\end{tabular}
}
\caption{\label{table_ablation}
\small{Ablation studies on the WMT15 De$\Rightarrow$En task. BLEU and AL scores are averaged under different latency settings. ``- w/o $\beta$'': remove sentence-level weight in Equation~\ref{eq:objectives}}. ``- w/o $D_{j,i}$'': remove diagonal regularization in Equation~\ref{eq:token-weight}.
} \vspace{-0.5cm}
\end{table}
In the CBSiMT framework, we apply weighted prefix-to-prefix training based on sentence-level and token-level weights. Diagonal regularization acts on token-level weights to encourage the model to predict along the ideal READ/WRITE path, while sentence-level weight is designed to reduce the chance of the model being exposed to non-monotonic sentence pairs during training.
Therefore we conduct ablation studies to demonstrate the effectiveness of the proposed two components.
For convenient comparison, we choose three settings for inference: $l_{max}$=$9$, $\delta=\left\{0.5,2.0,7.0\right\}$, and report the averaged BLEU and AL scores.
It can be seen from Table~\ref{table_ablation} that removing sentence-level weight $\beta$ or diagonal regularization $D_{j,i}$ will lead to a decrease in the BLEU score and an increase in the AL value, and removing both yields worse performing results.
So we can conclude that the sentence-level weight and diagonal regularization are both effective. 




\begin{CJK*}{UTF8}{gbsn}
\begin{table*}[!t]
\centering
\renewcommand\arraystretch{0.7}
{\small
  \begin{tabular}{m{2cm}|m{12.95cm}}
  \toprule[1pt]
    {\bf Source-1} & I found myself becoming a little bit of a technophobe \\ \midrule
  \end{tabular}
  \begin{tabular}{m{2cm}|m{0.25cm}m{0.7cm}m{0.7cm}m{0.7cm}m{0.3cm}m{0.3cm}m{2.5cm}m{4.575cm}}
    \multirow{2}*{\bf Reference-1} &  I & find & myself & a bit & like & a & technophobe & ~ \\ 
    ~ & 我 & 发掘 & 自己 & 有点 & 像 & 个 & 技术恐惧者 & ~ \\ \midrule
  \end{tabular}
  \begin{tabular}{m{2cm}|m{0.25cm}m{0.7cm}m{0.7cm}m{1.1cm}m{0.8cm}m{0.7cm}m{0.7cm}m{1.7cm}m{0.3cm}m{1.2cm}m{0.6cm}}
    \multirow{2}*{\bf Wait-$k$-1} & I & find & myself & become & a bit & like & a & technophobe & \textcolor{red}{'s} & \textcolor{red}{a part} & ~ \\ 
    ~ &  我 & 发现 & 自己 & 成为了 & 有点 & 像是 & 一个 & 技术恐惧症 & \textcolor{red}{的} & \textcolor{red}{一部分} & ~  \\ \midrule
  \end{tabular}
  \begin{tabular}{m{2cm}|m{0.25cm}m{0.7cm}m{0.7cm}m{0.8cm}m{2.5cm}m{5.9cm}}
    \multirow{2}*{\bf CBSiMT-1} & I & find & myself & a bit & technophobia & ~ \\ 
    ~ &  我 & 发现 & 自己 & 有点 & 技术恐惧症 & ~  \\ \midrule[1pt]
  \end{tabular}
  
  \begin{tabular}{m{2cm}|m{12.95cm}}
    {\bf Source-2} &  And so we're bringing them back in a contemporary story for children \\ \midrule
  \end{tabular}
  \begin{tabular}{m{2cm}|m{0.65cm}m{0.65cm}m{0.3cm}m{0.65cm}m{1.2cm}m{0.85cm}m{0.3cm}m{1.1cm}m{1.4cm}m{2.09cm}}
    \multirow{2}*{\bf Reference-2} & so & we & will & them & take back & take to & and & childrens & contemporary & in story \\ 
    ~ & 因此 & 我们 & 将 & 她们 & 带回来 & 带到 & 和 & 孩子们 & 同时代的 & 故事里   \\ \midrule
  \end{tabular}
  \begin{tabular}{m{2cm}|m{0.65cm}m{0.65cm}m{0.3cm}m{0.65cm}m{1.2cm}m{1.20cm}m{1.6cm}m{0.65cm}m{0.25cm}m{1.0cm}m{0.65cm}}
    \multirow{2}*{\bf Wait-$k$-2} & So & we & will & them & take back & \textcolor{red}{laboratory} & contemporary & story & \textcolor{red}{let} & childrens & \textcolor{red}{know} \\ 
    ~ &  所以 & 我们 & 将 & 它们 & 带回 & \textcolor{red}{实验室} & 现代 & 故事 & \textcolor{red}{让} & 孩子们 & \textcolor{red}{了解} \\ \midrule
  \end{tabular}
  \begin{tabular}{m{2cm}|m{0.65cm}m{0.65cm}m{0.3cm}m{0.65cm}m{1.2cm}m{1.4cm}m{1.1cm}m{0.3cm}m{1.1cm}m{1.7cm}}
    \multirow{2}*{\bf CBSiMT-2} & So & we & ~ & them & take back & contemporary & in story & for & childrens & ~ \\ 
    ~ & 所以 & 我们 & 把 & 它们 & 带回到 & 当代的 & 故事中 & 给 & 孩子们 & ~ \\ \bottomrule[1pt]
  \end{tabular}
}
\caption{
\small{Two examples in the test set of the MuST-C En$\Rightarrow$Zh task. Translations generated by the wait-$k$ and CBSiMT systems are compared. Hallucination tokens are colored in red.}} 
\vspace{-0.5cm}
\label{table:case-study}
\end{table*}
\end{CJK*}

\subsection{Hallucination Analysis} \label{sec:hallucination-ratio}
\begin{figure}[!t]
	\centering
	\begin{minipage}[c]{0.235\textwidth}
		\centering
		\includegraphics[width=\textwidth]{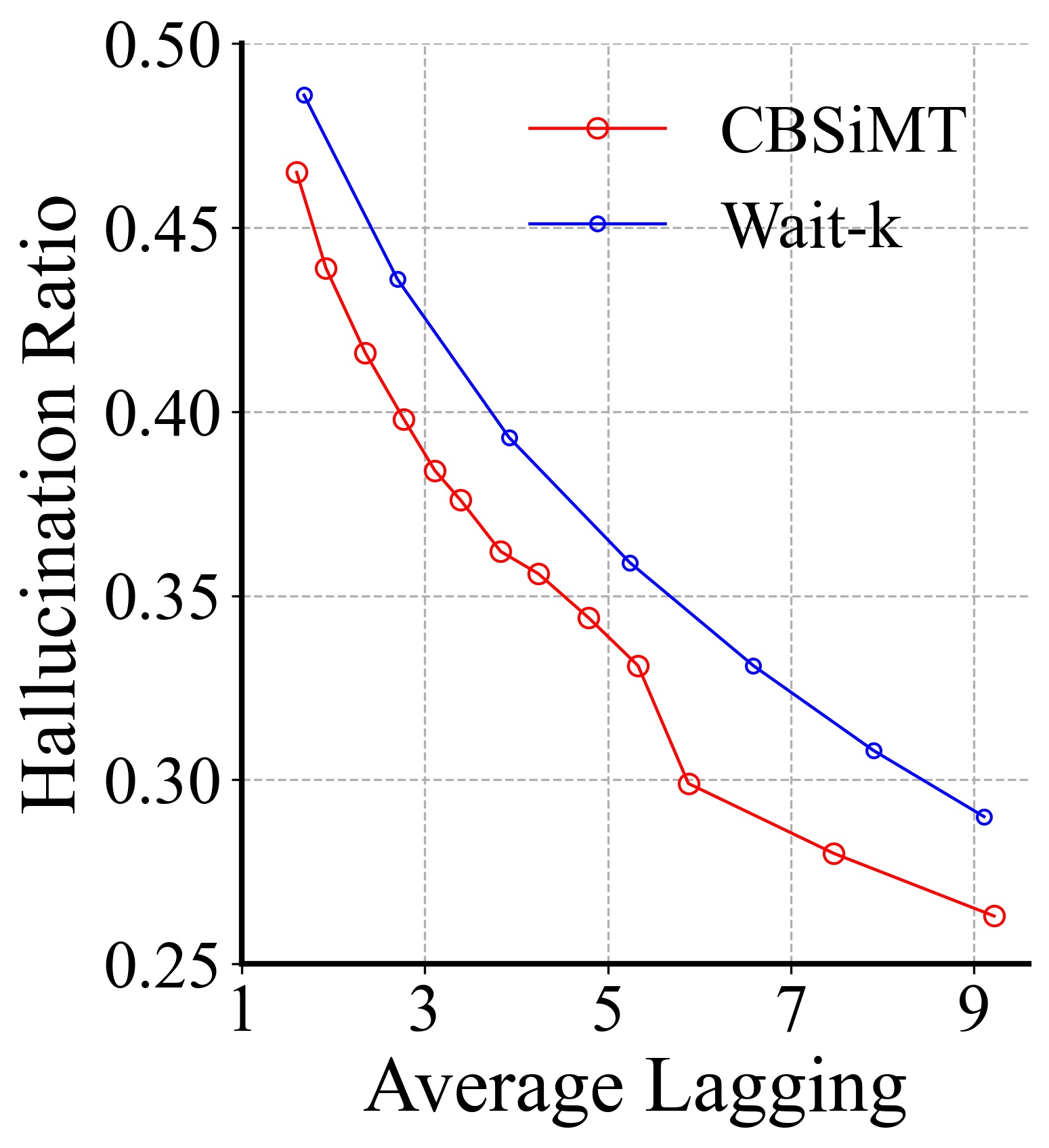}
		\subcaption{MuST-C En$\Rightarrow$Zh}
		\label{fig-mustc}
	\end{minipage}
	\begin{minipage}[c]{0.235\textwidth}
		\centering
		\includegraphics[width=\textwidth]{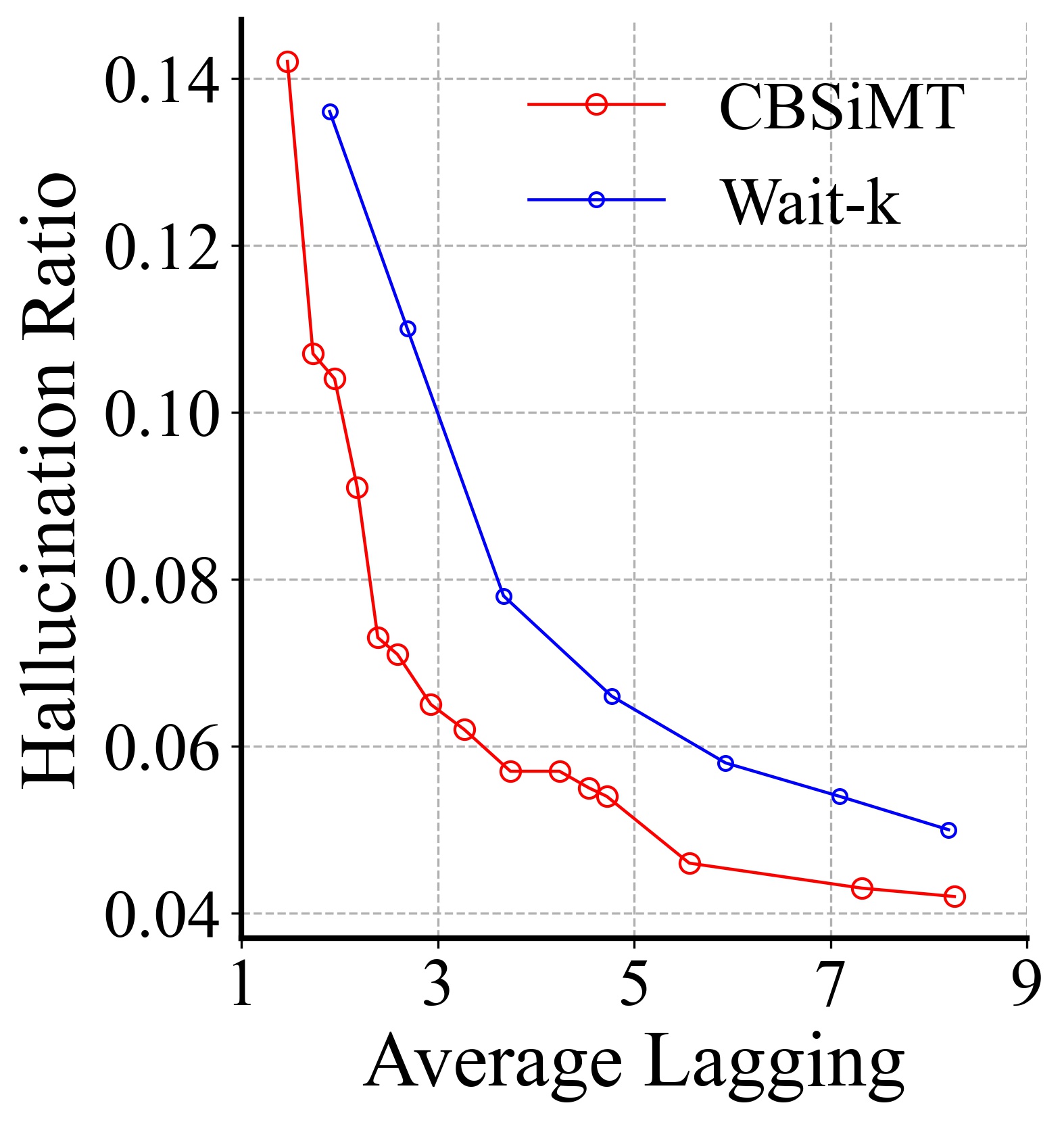}
		\subcaption{WMT15 De$\Rightarrow$En}
		\label{fig-wmt15}
	\end{minipage} 
	\caption{Comparison of hallucination rate versus AL value curves between CBSiMT and wait-$k$ systems on MuST En$\Rightarrow$Zh and WMT15 De$\Rightarrow$En tasks.}
	\label{figs:hallu-ratio} \vspace{-0.5cm}
\end{figure}

In this part, we investigate whether the CBSiMT framework can enhance the SiMT model by mitigating the negative impact of hallucination tokens on the model via weighted prefix-to-prefix training.
To identify hallucination tokens during inference, we record all prefix pairs corresponding to WRITE actions, and acquire the alignments of all prefix pairs with tool~\texttt{SimAlign}~\cite{jalili-sabet-etal-2020-simalign1}. Target tokens that are not aligned with the source prefix are considered hallucination tokens. The hallucination ratio is calculated as:
$$R_{H}=\frac{n_{t_H}}{n_t},$$
where $n_t$ is the number of all tokens in all translation prefixes, and $n_{t_H}$ represents the number of all hallucination tokens in all translation prefixes.

For CBSiMT and wait-k, we calculate the $R_{H}$ values corresponding to the translations under each latency setting. Hallucination-AL curves on two tasks are shown in Figure~\ref{figs:hallu-ratio}.
Since reading more source tokens leads to more accurate predictions from SiMT models, hallucination rates drop substantially for all systems as latency increases. On both tasks, CBSiMT has significantly lower hallucination ratios at each latency setting compared to wait-$k$. This supports our claim that CBSiMT can mitigate hallucinations in SiMT models.

\subsection{Case Study}
\label{sec:case_study}
\begin{CJK*}{UTF8}{gbsn}
We compare translations from different SiMT systems and observe hallucinations in translations. Translations for wait-$k$ and CBSiMT systems with similar latencies (note that both systems have an AL score of around $2.6$) are shown in Table~\ref{table:case-study}.
For the first example, the wait-$k$ system generates a long translation with hallucination tokens ``的一部分~(a part of)'' at the end.
For the second example, the wait-$k$ system produces a fluent prefix translation ``所以 我们 将 它们 带回 实验室~(So we take them back to the laboratory)'', but it contains a hallucination token ``实验室~(laboratory)'' whose information is not included in the source sentence, which further misleads the subsequent predictions.
Surprisingly, the two translations generated by CBSiMT have no hallucination tokens, effectively avoiding the hallucination phenomenon.
\end{CJK*}

\section{Related Work}
Works in SiMT generally follow two lines according to the inference policy: fixed policies and adaptive policies. We also involve the work of focusing on hallucinations in SiMT.
\paragraph{Fixed Policies.} Fixed policies employ predetermined rules to make READ/WRITE decisions, i.e., wait-$k$~\cite{ma2018stacl,elbayad2020efficient,zhang-feng-2021-universal1,zhang2021future}, which uses a simple and effective policy. But the fixed policy is not flexible enough for different inputs. 
\paragraph{Adaptive Policies.} Adaptive policies make dynamic decisions based on the context, such as MU~\cite{zhang-etal-2020-learning-adaptive1,zhang2022learning1}, MMA~\cite{ma2020monotonic,zhang2022gaussian,zhang2022modeling,zhang2022reducing}, ITST~\cite{zhang2022information}, and GSiMT~\citep{miao2021generative}. However, these methods usually require additional modules for READ/WRITE decisions, increasing model complexity. 
\paragraph{Hallucination.} 
Hallucination will produce pathological translations and is problematic for full-sentence NMT~\cite{lee2018hallucinations,muller-etal-2020-domain1,wang-sennrich-2020-exposure,raunak-etal-2021-curious,zhou-etal-2021-detecting}. SiMT is more susceptible to hallucination phenomena due to the limited source information.
~\citet{chen2020improving} generate monotonic pseudo data to improve performance and reduce hallucination. But it is costly to generate large-scale and high-quality monotonic pseudo data. 
~\citet{chang2022anticipation} apply a monotonic translation model to reduce anticipation and hallucination, yet the monotonic-translation step may cause translation quality to decrease.
~\citet{Han:Carpuat:Boyd-Graber-2022} utilize word-by-word question-answering evaluation tasks to reveal hallucination or omitting facts in SiMT systems.

\section{Conclusion}
In this work, we first demonstrate the hallucination caused by non-monotonic sentence pairs in SiMT models.
We further propose a CBSiMT framework to mitigate hallucination in SiMT.
Extensive experiments and in-depth analysis show the CBSiMT framework can significantly mitigate hallucinations and improve translation quality.

\section*{Limitations}
As mentioned above, the CBSiMT framework can significantly improve the performance of SiMT models on MuST English$\Rightarrow$Chinese and WMT15 German$\Rightarrow$English tasks.
However, the hallucination issues may be more serious for the translation tasks where the word order differs more (i.e., English$\Rightarrow$Japanese). Therefore, in future work, we plan to explore the performance of our proposed framework on more translation directions.

\section*{Ethics Statement}
This work aims to mitigate hallucination in simultaneous machine translation with weighted prefix-to-prefix training. In our experiments, all datasets are publicly available and commonly used in the SiMT community. Besides, all implementations are based on a widely-used open-source tool~$\texttt{fairseq}$~\cite{fairseq-cite}. The comparisons in this work are conducted based on the same experimental settings and datasets.


\bibliography{custom}
\bibliographystyle{acl_natbib}

\appendix



\end{document}